\documentclass[journal]{IEEEtran}
\usepackage{amsmath,amsfonts,amsthm}

\usepackage{algorithm2e}
\RestyleAlgo{ruled}

\usepackage{color}

\usepackage{float}
\usepackage{array}
\usepackage[caption=false,font=normalsize,labelfont=sf,textfont=sf]{subfig}
\usepackage{textcomp}
\usepackage{stfloats}
\usepackage{url}
\usepackage{verbatim}
\usepackage{graphicx}
\usepackage{cite}
\newtheorem{definition}{Definition}

\newtheorem{theorem}{Theorem}
\newtheorem{lemma}{Lemma}

\def\E{\mathbb{E}}
\def\P{\mathbb{P}}
\newcommand{\tr}{\mathrm{tr}}
\newcommand{\mv}{\Delta\mathbf{v}}  
\newcommand{\premu}{\boldsymbol{\mu}_0}
\newcommand{\presigma}{\boldsymbol{\Sigma}_0}
\newcommand{\postmu}{\boldsymbol{\mu}_1}
\newcommand{\postsigma}{\boldsymbol{\Sigma}_1}
\newcommand{\hpostmu}{\widehat{\boldsymbol{\mu}}_1}
\newcommand{\hpostsigma}{\widehat{\boldsymbol{\Sigma}}_1}
\newcommand{\lfun}{L}
\newcommand{\eth}{^{(e)}}
\newcommand{\epth}{^{(e+1)}}
\newcommand{\BerF}{\Phi}
\newcommand{\Berf}{\dot{\Phi}}
\newcommand{\mlog}{\log}
\newcommand{\mexp}{\exp}
\newcommand{\Lexp}{K_{\text{exp}}}
\newcommand{\Llog}{K_{\text{log}}}

\newcommand{\para}{\boldsymbol{\theta}}
\newcommand{\grid}{\mathcal{G}}
\newcommand{\branch}{\mathcal{E}}
\newcommand{\admit}{\boldsymbol{Y}_\grid}
\newcommand{\admitx}{\widetilde{\boldsymbol{Y}}_\grid}
\newcommand{\imped}{\boldsymbol{Z}_\grid}
\newcommand{\impedx}{\widetilde{\boldsymbol{Z}}_\grid}
\newcommand{\admitt}{\boldsymbol{Y}_\branch}
\newcommand{\admits}{\boldsymbol{Y}_\grid^s}
\newcommand{\incidence}{\boldsymbol{A}_{\branch, \grid}}
\newcommand{\incidencex}{\widetilde{\boldsymbol{A}}_{\branch, \grid}}
\newcommand{\rank}{\text{rank}}

\usepackage{tikz}
\usepackage{booktabs}
\usepackage{multirow}
\usepackage{enumitem}

\begin{document}

\title{Distribution Grid Line Outage Identification with Unknown Pattern and Performance Guarantee}

\author{Chenhan Xiao,~\IEEEmembership{Student Member,~IEEE}, Yizheng Liao,~\IEEEmembership{Member,~IEEE}, Yang Weng,~\IEEEmembership{Senior Member,~IEEE}
\vspace{-3em}
}



\maketitle

\begin{abstract}
Line outage identification in distribution grids is essential for sustainable grid operation. In this work, we propose a practical yet robust detection approach that utilizes only readily available voltage magnitudes, eliminating the need for costly phase angles or power flow data.
Given the sensor data, many existing detection methods based on change-point detection require prior knowledge of outage patterns, which are unknown for real-world outage scenarios. To remove this impractical requirement, we propose a data-driven method to learn the parameters of the post-outage distribution through gradient descent. However, directly using gradient descent presents feasibility issues. To address this, we modify our approach by adding a Bregman divergence constraint to control the trajectory of the parameter updates, which eliminates the feasibility problems.
As timely operation is the key nowadays, we prove that the optimal parameters can be learned with convergence guarantees via leveraging the statistical and physical properties of voltage data. We evaluate our approach using many representative distribution grids and real load profiles with 17 outage configurations. The results show that we can detect and localize the outage in timely manner with only voltage magnitudes and without assuming the prior knowledge of outage patterns.
\vspace{-0.5em}
\end{abstract}


\vspace{-1.0em}
\section{Introduction}
\vspace{-0.5em}
\label{sec:intro}
Distribution grid line outage occurrence detection and localization is essential for efficient system monitoring and sustainable system operation \cite{samudrala2020distributed}. A timely identification of the line outage effectively reduces potential financial loss. According to the U.S. Energy Information Administration, customers had an average of 1.3 outages and went without power for four hours during 2016 \cite{USreport}. The frequency and severity of line outages caused by extreme weather events and power supply shortages have also increased in recent years. 

The traditional line outage identification in distribution grids relies on passive feedback from customer reporting \cite{liao2021quick} or the ``last gasp'' message from smart meters \cite{bakken2014smart}, which is a notification automatically transmitted to the utility when power to the meter is lost. However, the performance of these methods will degrade while the transmission of the ``last gasp'' signal is not assured \cite{meier2019using}.
For instance, as the growth of distributed energy resources (DERs) penetration in distribution grids, customer can still receive power from the rooftop solar panels, battery storage, and electrical vehicles when there is no power flow in the distribution circuit connecting to the customer. So the smart meter at the customer premises cannot report a power outage.
Moreover, some secondary distribution grids are mesh networks in urban areas. In this scenario, a single line outage caused by circuit faults and human activities may not cause a power outage due to alternative paths for power supply. In this second case, we will also observe smart meters measuring power injections without sending the ``last gasp'' notification for reporting outages.

While alternative power sources make the ``last gasp'' notification fail to report outages, can we still find the line outage time and location? Answering this question, recent literature aimed at collecting additional information for smarter decisions. For example, power measurements, such as phasor angles from phasor measurement units (PMUs), were modeled in \cite{he2010fault} as a Gaussian Markov random field to track the grid topology change. Other power measurements, like power flows and load estimates, were also utilized in a compressive system \cite{babakmehr2019compressive} and hypothesis-test-based detection method \cite{sevlian2017outage}. Non-power measurements were explored as well, such as human network information from social media \cite{sun2016data} and the weather information from environment \cite{kankanala2013adaboost}. 
In distribution grids, obtaining measurements such as micro-PMUs and accurate power flow data can be challenging and costly, as they are not commonly deployed in households. To address this limitation, our earlier research \cite{liao2021quick} demonstrated that utilizing readily available voltage magnitudes could still yield accurate outage identification outcomes.
However, an in-depth examination of the probability distribution of voltage data and a theoretical guarantee for learning this distribution were not included in our previous work. These aspects are crucial for understanding the outage identification procedure. Besides, the method in \cite{liao2021quick} has feasibility and accuracy issues when learning the probability distribution. In this work, we fill the above gaps via a novel approach with theoretical guarantees.

To utilize the aforementioned measurements, both deterministic and probabilistic methods were proposed. Deterministic methods usually set up a threshold and declared the outage when the change of data exceeds the threshold. Such methods are simple to apply but cannot accurately discern data change in complex or large-scale grids. Probabilistic methods analyzed the data spatially or temporally. For spatial analysis, \cite{dwivedi2021scalable} studied graph spectral to assess the grid topology for line outage detection. However, such methods required the grid topology as a prior. For temporal analysis, tracing the probability distribution change of the time-series measurements is a common approach \cite{liao2021quick}. This is usually studied in the change point detection framework, which aims to find the distribution change of measurements as quickly as possible under the constraint of false alarm tolerance \cite{shiryaev1963optimum}. Such framework has been used in line outage and fault detection in transmission grids \cite{6250460,7041234} and DC micro-grids \cite{gajula2021quickest}. 
Although the change point detection framework assures optimal performance \cite{tartakovsky2005general}, it typically necessitates knowledge of both distributions before and after the change. Nevertheless, in distribution grids, this requirement is not practical as the post-outage distribution is unpredictable due to the large number of possible outage patterns, whereas the pre-outage distribution can be learned from historical measurements.

For removing the impractical requirement discussed above, methods were proposed to provide approximation or simplification of the unknown post-change distribution in change point detection. For instance, an approximated maximum likelihood estimation of unknown distribution parameters was proposed in \cite{liao2021quick}. A convexified estimation of the unknown distribution approach was introduced in \cite{cao2018sequential}. \cite{siegmund1995using, fotopoulos2010exact} bypassed the requirement in restricted distribution cases with partially unknown information (e.g., scalar Gaussian with unknown means and known variances). While these methods may mitigate the incompleteness of post-outage information, they have limitations on detection performance and parameter estimation.

In this paper, we propose a practical and straightforward method for utilities to identify line outages with unknown outage patterns. To address the challenge of limited data availability, our approach relies solely on voltage magnitudes obtained from smart meters. This is advantageous compared to expensive phase angle measurements and accurate power flow data, as voltage magnitudes are more readily accessible in typical distribution grids \cite{duan2021smart}. For the utilize of voltage magnitudes, we have made distinctive contributions. We demonstrate that the increment of voltage magnitudes before and after a line outage follows two distinct multivariate Gaussian distributions, where the distribution parameters are influenced by grid connectivity. Moreover, we provide theoretical guarantees for learning the unknown probability distribution parameters based on voltage magnitude data. By effectively utilizing voltage magnitudes and incorporating theoretical guarantees, we address the limitations posed by the absence of precise phase-angle data. Through the detection of changes in the learned Gaussian distributions, we can successfully identify line outages.

The second challenge is the unavailability of post-outage distribution parameters as analyzed earlier. To address this issue, we propose a data-driven method that directly learns these unknown parameters using Projected Gradient Descent (PGD). While Gradient Descent (GD) is susceptible to feasibility issues in parameter estimation, the iterative nature of GD allows us to control the parameter updating trajectory. Specifically, we formulate the distribution parameter learning problem as a projection optimization problem constrained by the Bergman divergence \cite{bregman1965finding}. This not only resolves the feasibility issue but also leads to accurate parameter estimation with theoretical guarantees. By accurately learning the parameters, our approach can effectively detect and localize line outages, even in large grids.

In addition to accuracy, utilities are also concerned with timely operation. By utilizing the statistical and physical characteristics of voltage data, we can limit the search space of unknown parameters to a convex set, which allows for fast and accurate recovery of the post-outage distribution. 
We have demonstrated that PGD can achieve optimal parameter learning with a polynomial-time convergence guarantee. Furthermore, we have developed an efficient implementation of the PGD algorithm, which reduces computational time by 75\% and makes it particularly well-suited for timely grid operations.

In summary, our proposed method offers several contributions. Firstly, it only requires simple data but have theoretical guarantees. Secondly, it does not require prior knowledge of the outage pattern. Thirdly, it enables timely operation. Furthermore, our approach comes with performance guarantees and does not rely on knowledge of the distribution grid's topology, nor does it require all households to have smart meters data.
The method is validated using four distribution grids and real-world load profiles with 17 outage configurations. In the following, Section \ref{sec:model} models the problem of line outage identification. Section \ref{sec:outage_detect} discusses the voltage data and identification procedure. Section \ref{sec:unknown+PGD} extends to identification with unknown outage pattern. Section \ref{sec:timely} provides performance guarantees on timely operation. Section \ref{sec:simulation} evaluates our method. Section \ref{sec:conclusion} concludes the paper.

\section{System Model}
\label{sec:model}

For showing our probabilistic design for change point detection and localization, we define variables on a graph probabilistically. Specifically, we model the distribution grid as a graph $\grid:=\{1,2,\cdots,M\}$ containing $M>0$ buses connected by branches. Then, the voltage data from each bus $i\in\grid$ is modeled as a random variable $V_i$. As a time-series, its realization at time $n$ is denoted as $v_i[n] = |v_i[n]| \exp( {j\theta_i[n]}) \in \mathcal{C}$, where $|v_i[n]| \in \mathcal{R}$ represents the voltage magnitude in per unit and $\theta_i[n] \in \mathcal{R}$ is the voltage phase angle in degrees. These steady-state measurements are sinusoidal signals at the same frequency. 
It's worth noting that unlike PMUs, smart meters typically do not measure phase angles. Therefore, we want to emphasize that even though voltage is represented in its phasor form, solely using the voltage magnitude can still effectively identify a line outage.

In the distribution grid $\grid$, the collection of voltage variables is modeled as $\mathbf{V}_\grid:=\left[V_1,V_2,\cdots,V_M\right]^\top\in\mathcal{R}^M$. Moreover, since $\mathbf{V}_\grid$ usually do not follow a regular distribution \cite{liao2021quick}, we model the increment change of voltage data as $\Delta \mathbf{V}_{\grid}$, whose realization at time $n$ is $\mv[n] = \mathbf{v}[n] - \mathbf{v}[n-1]$. For the sake of simplicity, we also use the notation $\mv^{1:N} = \{\mv[1],\cdots,\mv[N]\}$ to represent observations up to time $N$.

Based on the modeling, the problem of identifying the distribution grid line outage is formally defined as follows.
\begin{itemize}[leftmargin=*]
    \item {\bf Given}: Voltage increments $\mv^{1:N}$ from the smart meters.
    \item {\bf Find}: The line outage time as soon as possible and the out-of-service branch as accurate as possible.
\end{itemize}

\vspace{-0.5em}
\section{Outage Identification via Voltage Magnitude}
\label{sec:outage_detect}
While the expensive phasor angles and accurate power flows are hard to obtain in distribution grids, \cite{liao2021quick} showed that the easier-to-acquire voltage magnitude could be utilized to identify the line outage. The authors found that although voltage data do not follow a regular distribution, the incremental change of voltage follows Gaussian distribution. However, two things were missing in \cite{liao2021quick}: a clear formula of the distribution and an elaborate analysis of how such distribution is affected by line outages. They are the key to understanding the procedure and performance of identifying the line outage, which will be discussed in detail in the following subsection.

\vspace{-0.5em}
\subsection{Gaussian Distribution of Voltage Increment}

For answering the missing question, in this subsection, we elaborately prove that the increment of voltage data $\Delta\mathbf{V}_\grid$ follows two multivariate Gaussian distributions before and after the line outage, and provide a clear formula of such distribution. In doing so, we can identify the outage via tracing the change of the Gaussian distribution.

To study the distribution of $\Delta\mathbf{V}_\grid$, we start from the Kirchhoff's Current Law: the relationship between voltages $\mathbf{V}_{\grid}\in\mathcal{C}^{M}$ and currents $\mathbf{I}_{\grid}\in\mathcal{C}^{M}$ in the grid is $\admit \mathbf{V}_{\grid}=\mathbf{I}_{\grid}$, where the admittance matrix $\admit\in\mathcal{C}^{M \times M}$ can be derived through the connectivity of the grid as \cite{kettner2017properties}
\vspace{-0.3em}
\begin{equation}\label{eq:admit}
\admit = \incidence^\top \admitt \incidence + \admits.
\end{equation}
\vspace{-0.7em}

In the above equation, $\branch$ denotes the set of branches in the grid $\grid$. $\incidence\in\mathcal{R}^{|\branch|\times M}$ is the incidence matrix where each row represents a branch, and has exactly one entry of $1$ and one entry of $-1$ to denote the two buses connected by this branch. We can swap the $-1$ and $1$ since the grid network is undirectional.  $\admitt\in\mathcal{C}^{|\branch|\times |\branch|}$ is a diagonal matrix with the series admittances of each branch, and $\admits\in\mathcal{C}^{M\times M}$ is a diagonal matrix with the total shunt admittances at each bus.

By representing $\admit$ in \eqref{eq:admit}, we can discuss the invertibility of $\admit$, which prepares us for the distribution analysis of $\Delta\mathbf{V}_\grid$. To this end, we assume that the branches are not electromagnetically coupled and have non-zero admittance, i.e., $\admitt$ is full-rank. This assumption is common in distribution grids \cite{kettner2017properties}. With full-rank $\admitt$, we show the invertibility of $\admit$ in Lemma \ref{lemma:admit}.

\begin{figure*}[h]
\centering
\includegraphics[width=1\linewidth]{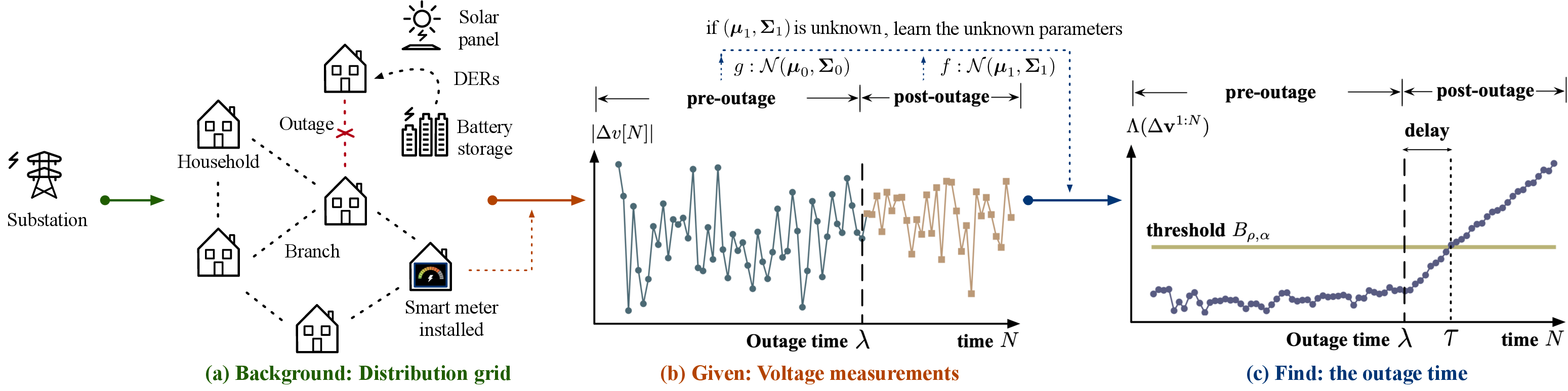}
\caption{An overview of the distribution grid line outage detection problem: we collect voltage magnitudes from smart meters installed at households and use the posterior probability ratio computed in \eqref{eq:posterior} to detect the change in the underlying distribution of voltage increments.}
\vskip -0.1in
\label{fig:bigpic}
\end{figure*}

\begin{lemma}\label{lemma:admit}
In a connected distribution grid $(\grid,\branch)$, the admittance matrix $\admit\in\mathcal{C}^{M \times M}$ is invertible after eliminating the slack-bus corresponding column and row.
\end{lemma}

In the following, we consider the eliminated admittance matrix and keep the notation unchanged for convenience. Based on Lemma $\ref{lemma:admit}$, the relationship between voltage increments $\Delta \mathbf{V}_{\grid}$ and current increments $\Delta \mathbf{I}_{\grid}$ can be expressed as 
\begin{equation}\label{eq:VandI}
\Delta \mathbf{V}_{\grid} = \imped \Delta \mathbf{I}_{\grid}, \quad \text{where} \quad \imped=\admit^{-1}.
\end{equation}

To derive the distribution of $\Delta \mathbf{V}_{\grid}$, we further introduce a common assumption regarding $\Delta \mathbf{I}_{\grid}$. 
We consider that $\Delta I$ at each non-slack bus is independent and normally distributed:
\begin{equation}\label{eq:assumptionI}
\Delta I_i\bot \Delta I_k, \  i\neq k \quad \text{and} \quad \Delta I_k\sim\mathcal{N}(\mu_k,\sigma^2_k), \  k\in\grid.
\end{equation}

\begin{figure}[H]
\centering
\vskip -0.10in
\includegraphics[width=0.7\linewidth]{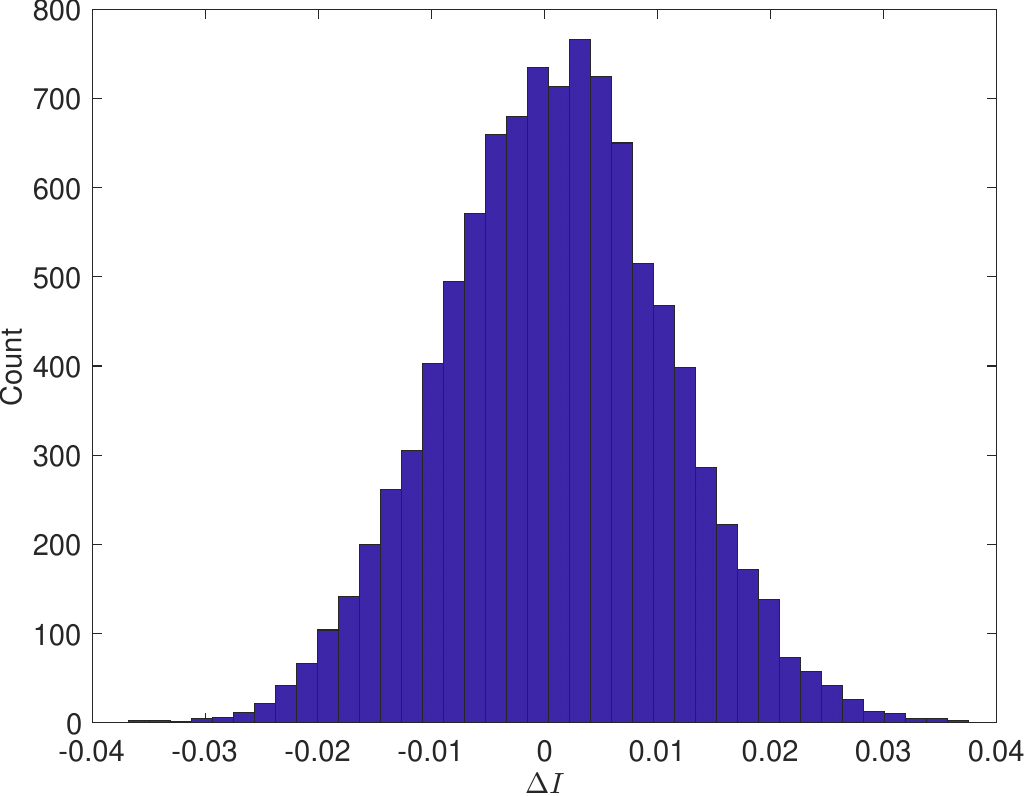}
\vskip -0.1in
\caption{The empirical histogram of $|\Delta I|$.}
\vskip -0.15in
\label{fig:MV}
\end{figure}

This statement is adopted and validated by real data in many works \cite{deka2017structure,bolognani2013identification, liao2018urban}, where the authors computed the mutual information between current injections to justify the independence. The empirical histogram in Fig. \ref{fig:MV} also suggests that $|\Delta I|$ roughly follows a Gaussian distribution. With the assumption in \eqref{eq:assumptionI}, we present the distribution analysis of $\Delta \mathbf{V}_{\grid}$, which is key to identifying the line outage.

\begin{theorem}\label{thm:multivariate} Provided that \eqref{eq:assumptionI} hold, $\Delta \mathbf{V}_{\grid}$ (excluding slack-bus) in a connected distribution grid $(\grid,\branch)$

\noindent 1) follows a multivariate Gaussian distribution,

\noindent 2) still follows a multivariate Gaussian distribution (with different mean and covariance) after grid topology changes.
\end{theorem}
\vspace{-0.5em}

\begin{proof}
With \eqref{eq:VandI}, $\Delta V_i$ of each bus $i\in\grid$ (slack-bus is excluded from $\grid$) can be expressed by $\Delta V_i = {\textstyle\sum}_{k\in\grid} Z_{ik} \Delta I_k$, where $Z_{ik}$ is the $(i,k)$ element of $\imped$. Hence, any non-trivial linear combination of $\Delta V_i,i\in\grid$ can also be represented by a linear combination of $\Delta I_k,k\in\grid$, and is normally distributed. This implies that the joint distribution of $\Delta \mathbf{V}_{\grid}=[\Delta V_1,\cdots,\Delta V_M]^\top$ is a multivariate Gaussian distribution as
\vspace{-0.2em}
\begin{equation}
    \Delta \mathbf{V}_{\grid} \sim \mathcal{N}(\boldsymbol{\mu}, \boldsymbol{\Sigma}),
\end{equation}
where $\boldsymbol{\mu}_i=\sum_{k\in\grid} Z_{ik}\mu_k$ and $\boldsymbol{\Sigma}_{ik}=\sum_{l\in\grid}Z_{il}Z_{kl}\sigma_l^2$.

When the grid topology is changed (e.g., due to a line outage), the incidence matrix $\incidence$ changes accordingly: if branch $l$ connecting bus $i$ and $k$ is out-of-service, $(\incidence)_{l,i}$ and $(\incidence)_{l,k}$  become zero. Denoting the new incidence matrix as $\incidencex$, there are two scenarios:
\begin{itemize}[leftmargin=*]
    \item The grid network is still connected (which is our focus in this paper). In this case, the changed admittance matrix $\admitx=\incidencex^\top \admitt \incidencex + \admits$ is still invertible, which results in a varied $\impedx=\admitx^{-1}$. It implies that $\Delta \mathbf{V}_{\grid}$ still follows a multivariate Gaussian distribution, only with different mean $\widetilde{\boldsymbol{\mu}}$ and different covariance $\widetilde{\boldsymbol{\Sigma}}$ calculated according to $\impedx$. 
    \item The grid network is disconnected. In this case, we view the network as disjoint islands where each part is a connected sub-network, e.g., $\grid=\grid_{1}\cup\grid_{2},\grid_{1}\cap\grid_{2}=\emptyset$. By doing so, we can write the incidence matrix in block format, e.g., $\incidencex=(\begin{matrix}\widetilde{\boldsymbol{A}}_{\branch_1, \grid_1}&\boldsymbol{0}\\\boldsymbol{0}&\widetilde{\boldsymbol{A}}_{\branch_2, \grid_2}\end{matrix})$. According to the first case, voltage increments in each sub-network follow a multivariate Gaussian distribution, and so does their joint distribution. In this scenario, since some houses lose power connection and will have zero voltages, the outage time and location can be more easily found via our approach. \qedhere
\end{itemize} 
\end{proof}

Suppose the outage occurs at time $\lambda$, Theorem \ref{thm:multivariate} allows us to write the sequence of voltage increments as
\begin{equation}\label{eq:voltage}
    \left\{ \begin{aligned}
        &\mv[n] \stackrel{i.i.d}{\sim} g:\mathcal{N}(\premu, \presigma), \quad n=&1,2,\cdots,\lambda-1,\\
        &\mv[n] \stackrel{i.i.d}{\sim} f:\mathcal{N}(\postmu, \postsigma), \quad n=&\lambda,\lambda+1,\cdots,N,
    \end{aligned} \right.
\end{equation}
where $g$ denotes the pre-outage Gaussian distribution and $f$ denotes the post-outage Gaussian distribution. The mean vectors $\premu,\postmu$, along with covariance matrices $\presigma,\postsigma$, are the parameters of these distributions. In our work, the pre-outage parameters $\premu$ and $\presigma$ can be estimated using historical data during normal operation periods of the distribution grid \cite{liao2021quick}. The post-outage parameters $\postmu$ and $\postsigma$ are considered unknown to reflect real-world outage scenarios, since the outage pattern is unpredictable/unknown. To visualize the varying distribution of the sequence, we provide an illustration of $|\Delta v[N]|\in \mathcal{R}$ in Fig. \ref{fig:bigpic}(b).

\vspace{-1em}
\subsection{Outage Identification via Distribution Change}
Before proposing our novel solution to unknown outage pattern, we present the commonly used framework to find the outage time $\lambda$ and outage branch, given voltage data in \eqref{eq:voltage}. 

To identify the outage time $\lambda$, we conduct a sequential hypothesis test $\mathcal{H}_0:\lambda>N$ and $\mathcal{H}_1:\lambda\leq N$ at every time step $N$. As $N$ increases, the first time we reject the null hypothesis $\mathcal{H}_0$ determines the value of $\lambda$. To decide when to reject $\mathcal{H}_0$, we compute the posterior probability ratio at each time step $N$ as
\begin{align}\label{eq:posterior}
&\Lambda(\mv^{1:N})=\frac{\P(\lambda\leq N|\mv^{1:N})}{\P(\lambda>N|\mv^{1:N})} \nonumber \\
&= \frac{\sum_{k=1}^N\pi(k)\prod_{n=1}^{k-1}g(\mv[n])\prod_{n=k}^{N}f(\mv[n])}{\sum_{k=N+1}^{\infty}\pi(k)\prod_{n=1}^{N}g(\mv[n])},
\end{align}
where $\lambda\in\mathbb{N}$ is assumed to follow a prior distribution $\pi$. The posterior probability ratio in \eqref{eq:posterior} compares the probabilities of ``outage occurred ($\lambda\leq N$)'' and ``outage did not occur ($\lambda> N$)'' based on the historical measurements $\mv^{1:N}$. A larger posterior probability ratio indicates that ``outage occurred'' is more likely than ``outage did not occur.'' Therefore, we declare the outage when the ratio \eqref{eq:posterior} exceeds a predefined threshold.
By the Shiryaev-Roberts-Pollaks procedure \cite{shiryaev1963optimum,tartakovsky2005general}, the following threshold in Theorem \ref{theorm:Bayesian} optimally considers the trade-off between the false alarm and the detection delay.


\begin{theorem}\label{theorm:Bayesian}
(Line outage detection). When $\lambda$ follows a geometric prior $\text{Geo}(\rho)$, we declare the outage at the first time when posterior probability ratio $\Lambda(\mv^{1:N})$ surpasses the threshold $B_{\rho,\alpha}=(1-\alpha)/(\rho\alpha)$ as
\begin{equation}\label{equ:stop rule}
\tau=\inf\{N\in\mathbb{N}:\Lambda(\mv^{1:N})\geq B_{\rho,\alpha}\},
\end{equation}
where the false alarm rate $\mathbb{P}(\tau<\lambda)$ is upper bounded by maximal false alarm rate $\alpha$. As $\alpha\to0$, $\tau$ in \eqref{equ:stop rule} is asymptotically optimal for minimizing the average detection delay $\E[\tau-\lambda|\tau\geq\lambda]$ as
\begin{equation}
\begin{aligned}
\E[\tau-\lambda|\tau\geq\lambda] &= \frac{|\log\alpha|}{-\log(1-\rho)+D_{KL}(f||g)}\\
&= {\textstyle\inf}_{\P(\tau^{\ast}\leq\lambda)\leq\alpha}\E[\tau^{\ast}-\lambda|\tau^{\ast}\geq\lambda],
\end{aligned}
\end{equation}
where $D_{KL}(f||g)$ is the KL divergence between $f$ and $g$.
\end{theorem}

One notable feature of the detection procedure described above is its ability to function effectively without requiring knowledge of the grid topology. Additionally, it can handle non-Gaussian distributions for $f$ and $g$. As depicted in Fig. \ref{fig:bigpic}(c), we calculate the posterior probability ratio sequentially and identify the outage time when the ratio exceeds the threshold.

Once the line outage occurrence is detected, localizing the out-of-service branch is also critical for system recovery. In \cite{liao2021quick}, the authors proposed an accurate outage localization method by proving that the voltage increments of two disconnected buses are conditionally independent. They computed the conditional correlation of every possible pair of buses in the grid and checked if the value changes from non-zero to zero. This approach differs from the utilization of nodal electric circuit matrices \cite{jamali2017fault, gururajapathy2017fault} for estimating fault location, while our approach has also been effective (as shown in Section \ref{sec:local_experiment}) and capitalizes on the learned covariance matrix in scenarios where the post-outage distribution is unknown.

To estimate the conditional correlation between bus $i$ and bus $k$, the covariance matrix $\boldsymbol{\Sigma}$ is utilized. Let set $\mathcal{I}:=\{i,k\}$ and $ \mathcal{K}:=\grid\backslash\{i,k\}$, the covariance matrix is decomposed as $\boldsymbol{\Sigma} = \left[ \begin{matrix}\boldsymbol{\Sigma}_{\mathcal{I}\mathcal{I}} & \boldsymbol{\Sigma}_{\mathcal{I}\mathcal{K}}\\ \boldsymbol{\Sigma}_{\mathcal{I}\mathcal{K}}^\top & \boldsymbol{\Sigma}_{\mathcal{K}\mathcal{K}}\end{matrix}\right]$. Based on this, the conditional correlation $\rho_{ik}$ between bus $i$ and bus $k$ is
\begin{equation}\label{eq:conditional}
    \rho_{ik}(\boldsymbol{\Sigma}) = \frac{\boldsymbol{\Sigma}_{\mathcal{I}|\mathcal{K}}(1,2)}{\sqrt{\boldsymbol{\Sigma}_{\mathcal{I}|\mathcal{K}}(1,1)\boldsymbol{\Sigma}_{\mathcal{I}|\mathcal{K}}(2,2)}},
\end{equation}
where the conditional covariance is computed by the Schur complement \cite{boyd2004convex} as $\boldsymbol{\Sigma}_{\mathcal{I}|\mathcal{K}} = \boldsymbol{\Sigma}_{\mathcal{I}\mathcal{I}} - \boldsymbol{\Sigma}_{\mathcal{I}\mathcal{K}}\boldsymbol{\Sigma}_{\mathcal{K}\mathcal{K}}^{-1}\boldsymbol{\Sigma}_{\mathcal{I}\mathcal{K}}^\top$.

\begin{theorem}(Line outage localization). \label{lemma:local}
The conditional correlation is calculated based on \eqref{eq:conditional} for every pair of $(i,k)$ as \vspace{-0.5em}
\begin{equation}
    \underbrace{\rho_{ik}^{-} = \rho_{ik}(\presigma)}_{\text{before outage}}\quad\text{and}\quad \underbrace{\rho_{ik}^{+} = \rho_{ik}(\hpostsigma)
    }_{\text{after outage}}.\vspace{-0.5em}
\end{equation}
The branch between bus $i$ and $k$ is out-of-service if
$
|\rho_{ik}^{-}| > \delta_{\max}$ and $|\rho_{ik}^{+}| < \delta_{\min}$. The thresholds are set as $\delta_{\max}=0.5$ and $\delta_{\min}=0.1$ based on real-world outage data to check if the correlation changes from non-zero to near-zero value.
\end{theorem}
According to Theorem \ref{lemma:local}, we track the change of covariance matrices to localize the out-of-service branch. Specifically, an out-of-service branch between bus $i$ and bus $k$ can be identified if both of the following conditions are met simultaneously: (1) $|\rho_{ik}^{-}| > \delta_{\max}$ indicating the presence of a branch between buses $i$ and $k$ before the outage, and (2) $|\rho_{ik}^{+}| < \delta_{\min}$ indicating the absence of a branch between buses $i$ and $k$ after the outage.
Notably, this process still does not need the grid topology as a prior.

\section{Outage Identification with Unknown Pattern }
\label{sec:unknown+PGD}
The detection and localization procedure in Section~\ref{sec:outage_detect} requires knowing all the parameters of $g$ and $f$ in advance. However, this is impractical in real-world distribution grids. Specifically, although we know $g$ and $f$ are multivariate Gaussian distributions based on Theorem \ref{thm:multivariate}, the parameters (mean vectors and covariance matrices) of $f$ are usually hidden. While the pre-outage distribution parameters can be learned by historical measurements during the normal grid operation, the post-outage distribution parameters are often unavailable. In fact, since there are a large number of branches and therefore, substantial combinations of outage patterns, we can not predict the outage pattern or the post-outage distribution parameters. Hence, we need to estimate the unknown parameters before conducting the aforementioned methods to identify the outage.

To resolve such issue, we propose a data-driven framework to learn the post-outage distribution parameters $\para = (\postmu,\postsigma)$ jointly. Specifically, we want to find the parameter set that minimizes the negative likelihood function $\lfun(\postmu,\postsigma)$ as
\vspace{-0.1em}
\begin{align}\label{eq:optimization}
    (\hpostmu, \hpostsigma) &= \arg\min_{(\postmu,\postsigma)} \lfun(\postmu,\postsigma),
\end{align}
where $\lfun(\postmu,\postsigma)$ is computed as
\vspace{-0.1em}
\begin{align}\label{eq:likelihood}
-\sum_{k=1}^N\pi(k)\prod_{n=1}^{k-1}g(\mv[n])\prod_{n=k}^{N}f(\mv[n]|\postmu,\postsigma).
\end{align}

To address the non-convex nature of the likelihood expressed in equation \eqref{eq:likelihood}, the authors in \cite{liao2021quick} proposed a convex approximation using Jensen's inequality and derived closed-form solutions for equation \eqref{eq:optimization}. However, the use of Jensen's inequality can introduce inaccuracies in the resulting closed-form solutions, particularly in determining the minimum point. Furthermore, the estimated covariance matrix may not always be feasible. Specifically, a feasible covariance matrix must be positive definite, i.e., $\postsigma\succ0$, and if this condition is not met during the learning process, the computation of the probability density of $f$ can fail.

An alternative approach is using the Gradient Descent (GD) to find the solution to \eqref{eq:optimization}. While the vanilla GD also can not ensure the aforementioned feasibility of the parameters, the iterative learning nature in GD enables us to control the updating trajectory of parameters.

\vspace{-1.0em}
\subsection{Unknown Parameters Estimation via Projected Gradient Descent with Bregman Divergence Constraint}
To guarantee that the estimation of parameters $\para=\{\postmu,\postsigma\}$ are always feasible, we introduce the Bregman divergence \cite{bregman1965finding} to constrain the estimate in each iteration of GD and arrive at a series of optimization problems as
\begin{equation}\label{eq:bregman}
    \para_{i}\epth = \arg\min_{\para_{i}} \underbrace{\Delta_{\BerF} (\para_{i}, \para_{-i}\eth)}_{\text{Bregman divergence}} + \ \eta \lfun (\para_{i}, \para_{-i}\eth).
\end{equation}

In the above equation, $\para_i\eth$ is the update of the $i^{th}$ parameter at $e^{th}$ iteration, $\para_{-i}\eth=\para\eth \setminus \para_i\eth$ is a complement set, and $\eta$ is the trade-off learning rate. The Bregman divergence
\begin{align*}
    \Delta_\BerF(\para_{i}, \para_{i}\eth):= \BerF(\para_{i}) - \BerF(\para_{i}\eth) - \tr\left( (\para_{i}-\para_{i}\eth)\BerF(\para_{i}\eth)^\top\right)
\end{align*}
provides a distance measurement between two variables $\para_i$ and $\para_i\eth$, where $\BerF$ is a strictly convex differentiable function. Intuitively, it restricts the new estimate $\para_i\epth$ to stay relatively close to the previous estimate $\para_i\eth$. Therefore, if the initial guess $\para_i^{(0)}$ is within the feasible domain, we can expect that the parameter $\para_i$ following \eqref{eq:bregman} will update towards the direction of minimizing the negative likelihood and meanwhile satisfy a similar property (e.g., positive definiteness of covariance matrix) since the update step is restricted.

Finding the solution to \eqref{eq:bregman} relies on one characteristic of Bregman divergence: its gradient with respect to $\para_i$ has a simple form
$
    \nabla_{\para_{i}} \Delta_\BerF(\para_{i}, \para_{i}\eth) = \Berf(\para_{i}) - \Berf(\para_{i}\eth),
$
where $\Berf$ is the differential of $\BerF$. Based on this, we can eliminate the argmin in $\eqref{eq:bregman}$ by setting the gradient (with respect to $\para_i$) of the objective function to zero, and derive the following Projected Gradient Descent solution \cite{tsuda2004matrix}.

\begin{lemma} \label{lemma:PGD}
The optimization problem in \eqref{eq:bregman} is solved as
\begin{equation}\label{eq:PGD0}
    \para_i\epth = \Berf^{-1}\left(\Berf(\para_i\eth) + \eta \nabla_{\para_i}\lfun(\para\eth)\right).
\end{equation}
\end{lemma}

With Lemma \ref{lemma:PGD}, we propose the important result of our paper: the learning scheme of unknown parameters with feasibility guarantee. We will further show that this learning scheme is accurate and has convergence guarantees.

\begin{theorem} (Projected Gradient Descent of learning $\postmu, \postsigma$).\ \label{Theorem: PGD}
With careful customization of Bregman divergence (i.e., choosing the appropriate function $\BerF$), the Projected Gradient Descent learning in \eqref{eq:PGD0} becomes feasible.
\begin{itemize}[leftmargin=*]
    \item \underline{$\postmu \in \mathcal{R}^{M}$}: $\BerF(\postmu) = \frac{1}{2}\|\postmu \|_2^2$ and $\Berf^{-1}(\postmu )=\postmu $. The learning scheme is
    \begin{equation}
        \postmu\epth = \postmu\eth - \eta \nabla_{\postmu}\lfun(\postmu\eth,\postsigma\eth). \label{eq:PGD_mean}
    \end{equation}

    \item \underline{$\postsigma\succ0$}: $\BerF(\postsigma) = \tr(\postsigma\log\postsigma-\postsigma)$ and $\Berf^{-1}(\postsigma)=\exp \postsigma$. The learning scheme is
    \begin{equation}
        \postsigma\epth = \exp\left(\log\postsigma\eth - \eta \nabla_{\postsigma}\lfun(\postmu\eth,\postsigma\eth)\right). \label{eq:PGD_covariance}
    \end{equation}
\end{itemize}
\end{theorem}

\begin{figure}[ht]
\centering
\vskip -0.10in
\includegraphics[width=0.7\linewidth]{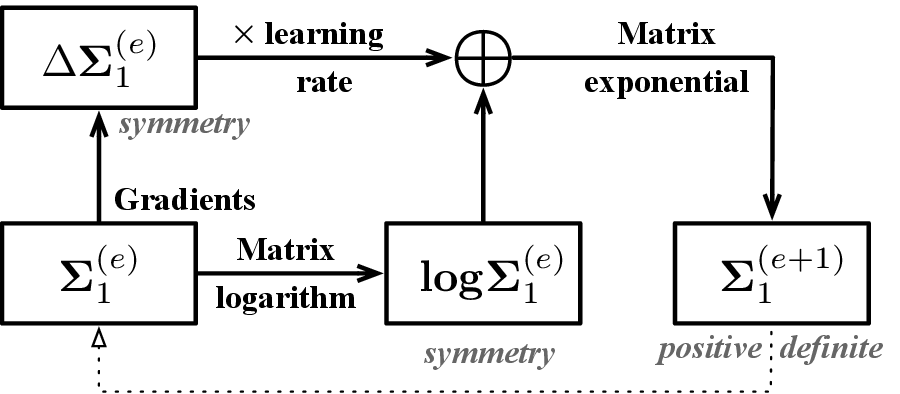}
\vskip -0.1in
\caption{The Projected Gradient Descent update of $\postsigma$.}
\vskip -0.15in
\label{fig:PGD}
\end{figure}

The learning process of $\postsigma$ is shown in Fig. \ref{fig:PGD}. Since matrix exponential maps any symmetric matrix to a positive definite matrix, the learning scheme \eqref{eq:PGD_covariance} maintains the property of positive definiteness, i.e., $\postsigma\epth\succ0$ if $\postsigma\eth\succ0$. 

Besides the statistical properties (e.g., the covariance matrix is positive definite), smart meter data have physical properties as well due to grid operation. For example, because the standard range of voltage magnitude is between $0\ p.u.$ and $1.1\ p.u.$, the mean value of voltage increment should be between $-1.1\ p.u.$ and $1.1\ p.u.$. Our learning scheme in \eqref{eq:PGD0} can also satisfy this requirement by defining 
\vspace{-0.7em}
\begin{align}
\BerF(\postmu) = \sum_{i=1}^M\left[(\mu_i+1.1)\log(\mu_i+1.1) \right. \nonumber \\
\left.+ (1.1-\mu_i)\log(1.1-\mu_i) + \mu_i\right], \label{eq:PGD_mean1}
\end{align}
where $\mu_i$ is the $i^{th}$ element of the mean vector $\postmu$. 

To conclude, when post-outage distribution parameters $\postmu,\postsigma$ are unknown, we use \eqref{eq:PGD_mean1} and \eqref{eq:PGD_covariance} to accurately learn them with feasibility and convergence guarantees. The pre-outage parameters $\premu$ and $\presigma$ can be estimated using historical data during normal operation periods of the distribution grid \cite{liao2021quick}. By obtaining these parameters of the Gaussian density functions $g$ and $f$ in \eqref{eq:posterior}, we can explicitly calculate $g$ and $f$. It enables us to implement Theorem \ref{theorm:Bayesian} for detecting the outage time and use Theorem \ref{lemma:local} for localizing the outage branch. This framework is summarized into Algorithm \ref{algo:PGD}.

\begin{algorithm}
\DontPrintSemicolon
\caption{Line outage identification with unknown post-outage distribution parameters}
\label{algo:PGD}
    \KwIn{New observation $\mv[N]$}
    \KwOut{Outage time $\tau$ and outage location}
    
        Set $\postmu^{(0)}, \postsigma^{(0)}$ from ${\scriptstyle (N-1)}^{th}$ step \tcp*{\footnotesize warm start}
        
        \For{$e=0,1,\ldots$}
        {
          $\postmu\epth \leftarrow \Berf^{-1}\left(\Berf(\postmu\eth)-\eta \nabla_{\postmu}\lfun\right)$ \tcp*{\footnotesize use $\BerF$ in \eqref{eq:PGD_mean1}}
          
          $\postsigma\epth \leftarrow \exp\left(\log\postsigma\eth-\eta\nabla_{\postsigma}\lfun\right)$
          
          \uIf{$|\lfun(\postmu\eth,\postsigma\eth) - \lfun(\postmu\epth,\postsigma\epth)|\le10^{-3}$}
          {
            \Return {$\postmu^{\text{best}},\postsigma^{\text{best}}$} \tcp*{\footnotesize return best update result, defined in Theorem \ref{theorem:mu_converge}}
          }
        }
    \uIf{$\Lambda{(}\mv^{1:N}{)} \geq B_{\rho,\alpha}$} 
    {
      \For{$i,k\in\grid$}
      {
        \uIf{$|\rho_{ik}^{-}| > \delta_{\max}$ and $|\rho_{ik}^{+}| < \delta_{\min}$}
        {
            \Return {$\tau = N$, report the out-of-service branch between bus $i$ and $k$}
        }
      }
    }
\end{algorithm}


\vspace{-1.5em}
\section{Timely Outage Identification with Performance Guarantee}
\label{sec:timely}
\vspace{-0.2em}
In addition to the feasibility issue that has already been addressed in Theorem \ref{Theorem: PGD}, the accuracy and computation time of the proposed learning scheme are two other concerns when implementing such a method for real-world outage identification. In this section, we demonstrate that our proposed method can achieve the optimal parameter solution with a guaranteed convergence. Furthermore, we present an efficient implementation for timely operation.

\vspace{-0.5em}
\subsection{Restricted Convexity for Convergence Guarantee}
\label{sec:convergence}

While the non-convexity of likelihood $\lfun(\postmu,\postsigma)$ in \eqref{eq:likelihood} hinders us from deriving a convergence analysis directly, we note that $\lfun(\postmu,\postsigma)$ is constrained convex. Specifically, we notice that the unknown parameters $\postmu,\postsigma$ of $f$ are not supposed to be far away from the known parameters $\premu,\presigma$ of $g$, thus freeing us from searching the entire parameter space. This is because the alternative power supply makes the impact of line outage less severe. In fact, if $\postmu,\postsigma$ are significantly far from $\premu,\presigma$, distinguishing between $f$ and $g$ would become trivial: imaging a large leap from pre-outage data to post-outage data in Fig.  \ref{fig:bigpic}(b), one can detect this change very easily. Hence, a reasonable assumption is that $\postmu,\postsigma$ are relatively close to $\premu,\presigma$, which actually results in a much harder detection problem. With this assumption, we can restrict our search for parameters in a constrained set where $\lfun(\postmu,\postsigma)$ has good properties. To formally present this, we introduce the restricted convexity \cite{jain2017non}.

\begin{definition}\label{def:RC}
A continuously differentiable function $H : \mathcal{R}^M \to \mathcal{R}$ is restricted convex over a possibly non-convex region $\mathcal{D}\subseteq \mathcal{R}^M$ if every $\boldsymbol{x},\boldsymbol{y}\in\mathcal{D}$ we have $H(\boldsymbol{y}) \ge H(\boldsymbol{x}) + \langle\nabla H(\boldsymbol{x}), \boldsymbol{y}-\boldsymbol{x}\rangle$.
\end{definition}
\vspace{-0.1em}
Then, we show that $\lfun(\postmu,\postsigma)$ satisfies the restricted convexity in Definition \ref{def:RC}. Specifically, $\lfun(\postmu)$ is restricted convex on constrained set $\left\{\postmu\left| \postsigma \succeq \frac{ \boldsymbol{v}_k(\postmu)\boldsymbol{v}_k(\postmu)^T}{(N-k+1)} , \forall k\le N \right. \right\}$ where $\boldsymbol{v}_k(\postmu)=\sum_{n=k}^{N}(\mv[n]-\postmu)$. Similarly, $\lfun(\postsigma)$ is restricted convex on constrained set $\{\postsigma\succ0|\nabla^2 \lfun({\rm vec}(\postsigma))\succeq 0\}$. Based on this property, we derive in Theorem \ref{theorem:mu_converge} the convergence of updating $\postmu$ and $\postsigma$.

\begin{figure}[H]
\begin{center}
\vskip -0.1in
\includegraphics[width=0.95\linewidth]{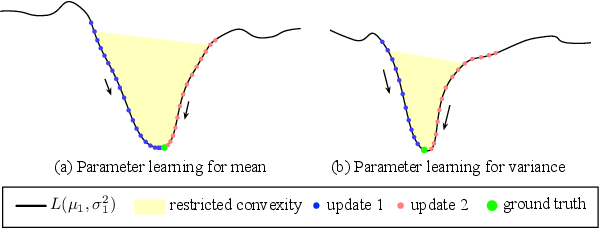}
\vskip -0.1in
\caption{Visualization of learning parameters in $g\sim\mathcal{N}(0,\frac{1}{2}),f\sim\mathcal{N}(1,\frac{1}{5})$.}
\vskip -0.2in
\label{fig:converge1}
\end{center}
\end{figure}

\begin{theorem}\label{theorem:mu_converge} Using PGD to iteratively update $\para_i$, the best update $\para_i^{\text{best}}:=\arg\min_{e\in[E]} \lfun(\para_i\eth)$ and the averaged update $\para_i^{\text{avg}}:=\frac{1}{E}\sum_{e=1}^E\para_i\eth$ will converge to the optimal value $\para_i^{\ast} = \arg\min_{\para_i}\lfun(\para_i)$ with a step size $\eta=\frac{1}{\sqrt{E}}$: 
\vspace{-0.1em}
$$
\lfun(\para_i^{\text{best}}) \le \lfun(\para_i^{\ast}) + \varepsilon\quad\text{and}\quad \lfun(\para_i^{\text{avg}}) \le \lfun(\para_i^{\ast}) + \varepsilon,
$$
for any $\varepsilon>0$, after at most $E=\mathcal{O}(\frac{1}{\varepsilon^2})$ iterations.
\end{theorem}
The proof is in Appendix \ref{app:proofconvergence}. Moreover, since the best update converges faster as shown in Section \ref{sec:simulation}, we choose it as the output of the learning scheme in Algorithm \ref{algo:PGD}. To better visualize how the parameters are updated in the restricted convex area via Projected Gradient Descent (PGD), we provide Fig. \ref{fig:converge1}. As we see, although the likelihood function $\lfun$ is not convex in the entire parameter space, the restricted area is convex, opening the door for learning accurate and feasible parameter solutions to \eqref{eq:optimization}.

\subsection{Acceleration for Timely Operation}
\label{sec:speedup}
Theorem \ref{theorem:mu_converge} shows that our proposed method can find the optimal parameters with polynomial-time complexity, which enables quick operation. In this subsection, we provide an efficient implementation of the learning scheme to further accelerate the algorithm for timely outage identification. To achieve so, we notice that while the matrix exponential and logarithm operations in \eqref{eq:PGD_covariance} provide good properties of covariance estimation, it is very time-consuming when calculating them. The calculation is time-consuming because it is often based on their infinite Taylor series. To accelerate these operations, we propose to use finite terms of their Taylor series to approximate the operations.

The matrix exponential is given by the power series in \eqref{eq:exp_app}, and can be approximated by its first $\Lexp$ terms since $\frac{1}{k!}$ decreases drastically when $k$ becomes large. However, once we replace the original matrix exponential operation with the approximated operation $\widehat{\mexp}$, we need to verify that $\widehat{\mexp}$ also maps any symmetric matrix to a positive definite matrix to satisfy the conclusion in Theorem \ref{Theorem: PGD}. Otherwise, we will arrive at covariance estimates outside the feasible domain. With this motivation, we show in Lemma \ref{lemma:exp_speedup} that if we choose an appropriate value of $\Lexp$ for approximation, the operation $\widehat{\mexp}$ has similar properties as $\mexp$.

\begin{lemma}\label{lemma:exp_speedup}
The matrix exponential can be approximated as
\begin{align}\label{eq:exp_app}
    \mexp (\boldsymbol{X}) &= {\textstyle\sum}_{k=0}^{\infty}\frac{1}{k!}\ \boldsymbol{X}^k \approx {\textstyle\sum}_{k=0}^{\Lexp} \frac{1}{k!}\ \boldsymbol{X}^k := \widehat{\mexp}(\boldsymbol{X}),
\end{align}
for any real matrix $\boldsymbol{X}$. Also, $\widehat{\mexp}(\boldsymbol{X})\succ 0$ for any symmetric $\boldsymbol{X}$ if $\Lexp$ is even and $\Lexp>\max\{0,-a_{\min}\}$, where $a_{\min}$ is the smallest eigenvalue of $\boldsymbol{X}$.
\end{lemma}

Similarly, the matrix logarithm is given by the power series in \eqref{eq:log_app}, and can be approximated by the first $\Llog$ terms. 

\begin{lemma}\label{lemma:log_speedup}
The matrix logarithm can be approximated as
\begin{align}\label{eq:log_app}
    \mlog (\boldsymbol{X}) 
    &\approx {\textstyle\sum}_{k=1}^{\Llog} \frac{1}{k} (-1)^{k+1} (\boldsymbol{X} - \boldsymbol{I})^k := \widehat{\mlog}(\boldsymbol{X}).
\end{align}
\end{lemma}

In order to make Theorem \ref{Theorem: PGD} still hold true when we use the approximated operation $\widehat{\mlog}$, we only need the symmetry of $\widehat{\mlog}$, which can be easily verified as $\widehat{\mlog} (\boldsymbol{X}^\top) = {\textstyle\sum}_{k=1}^{\Llog} \frac{1}{k} (-1)^{k+1} \left((\boldsymbol{X} - \boldsymbol{I})^k\right)^\top = \left(\widehat{\mlog} (\boldsymbol{X})\right)^\top.$

In summary, the proposed two approximation operations $\widehat{\mexp}$ and $\widehat{\mlog}$ can still preserve the feasibility of the covariance matrix estimation. The choice of $\Lexp$ and $\Llog$ is a trade-off between execution time and approximation accuracy: when $\Lexp$ is small, the operation $\widehat{\mexp}$ is very fast but provides a poor approximation of $\mexp$ and vice versa. In Section \ref{sec:simulation}, we will demonstrate how we choose an appropriate approximation level which results in almost zero errors and over $75\%$ reduction in execution time.

\vspace{-0.8em}
\section{Validate on Extensive Outage Scenarios with Real-World Data}
\label{sec:simulation}

This section shows how our proposed method performs in various distribution grids with real-world data. To evaluate our method in systems with different sizes and environments, we design extensive experiments on IEEE 8-bus, IEEE 123-bus networks \cite{kersting1991radial}, as well as two European representative distribution systems: medium voltage (MV) network in the urban area and low voltage (LV) network in the suburban area \cite{mateo2018european}. In each network, bus 1 is selected as the slack bus.

To account for more complex outage scenarios in real-world distribution grids, we examine situations where alternative power sources are available after a line outage. In these scenarios, the "last gasp" notification is ineffective, making it more difficult to detect the line outage. We simulated the following two representative scenarios to replicate this complex scenario. It should also be noted that if certain buses are disconnected from the main grid and experience a voltage magnitude of zero following an outage, our method can accurately and quickly identify the out-of-service line. This is a simpler case compared to the ones we simulated below.

\begin{itemize}[leftmargin=*]
    \item \underline{Mesh networks} where most buses have non-zero voltages after the outage since they can receive power from alternative branches. Mesh network often depicts the outage scenario in urban areas. For simulating mesh networks, we add loops in each aforementioned network to ensure it is still connected after line outages, following the study in \cite{liao2021quick} and \cite{liao2018urban}.
    
    \item \underline{Radial networks with DERs} where some buses still receive power from DERs though they are isolated from the main grid after an outage (see Fig. \ref{fig:bigpic}(a)). This outage scenario is typical in residential areas. To simulate DERs, we select multiple buses to have solar power panels with batteries as the storage. For the solar panel, we use the power generation profile computed by PVWatts Calculator \cite{dobos2014pvwatts}.
\end{itemize}

For simulating more realistic data, we use the real residential power profile from Duquesne Light Company (DLC) in Pittsburgh, PA, USA. This profile contains anonymized and secure hourly (and 15-minute) smart meter readings of active power over more than 5,000 houses in the year 2016. The basic statistics of this dataset are summarized in Table \ref{tab:DLC}.
\begin{table}[H]
    \centering
    \vskip -0.15in
    \caption{Statistical analysis of DLC power dataset.}
    \label{tab:DLC}
    \vskip -0.1in
    \begin{tabular}{c|c}
    \toprule
    Statistics & Value \\
    \midrule
    Minimum Value        & $-2.6040$    \\
    Maximum Value      & $26.6860$   \\
    Mean     & $0.8473$    \\
    Standard Deviation   & $0.6387$    \\
    Skewness & $1.7441$\\
    \midrule
    \end{tabular}
    \vskip -0.10in
\end{table}

The time-series voltage data are simulated by the MATLAB Power System Simulation Package (MATPOWER) in MATLAB R2022b.
In every distribution network, we assign active power $p_i[n]$ from the above DLC power profile to each bus $i$ at time $n$. The reactive power $q_i[n]$ is computed according to a randomly generated power factor $pf_i[n]$, which follows a uniform distribution, e.g. $pf_i[n]\sim \text{Unif}(0.9,1)$. Based on the active and reactive power, we use MATPOWER to solve power flow equations and obtain voltage measurements. Moreover, we can simulate an outage scenario by setting the admittance of a branch or several branches to zero. Hence, we can generate the voltage data during both normal operation and outage scenarios. 
After we obtain voltage data from MATLAB, all the remaining calculations in outage detection in Algorithm \ref{algo:PGD} are implemented with Python 3.8 on a personal computer with a Windows 10 operating system, an Intel Core i7 processor clocked at 2.2 GHz, and 16 GB of RAM.

Due to the limited deployment of PMU in reality, the voltage phase angles are hard to obtain. Hence, as mentioned in Section \ref{sec:model}, we only use the voltage magnitude in the following experiments even though we model the voltage data in its phasor form. Another concern of data is the high dimensionality in large-scale grids. To resolve this computational issue, we apply the whitening transformation to our data as $\mv^{1:N} \to \boldsymbol{W}\mv^{1:N}$ based on the PCA whitening matrix satisfying $\boldsymbol{W}^\top\boldsymbol{W}=\presigma^{-1}$. Since the whitening transformation does not change the KL divergence between $g$ and $f$, it has no impact on the outage detection performance.

In the subsequent experiments, we compare our proposed method with various baselines. When full knowledge of post-outage distribution $f$ is known, we refer to the optimal Bayesian procedure as {\bf $f$ known}. When the parameters of $f$ are unknown, our method is referred to as {\bf PGD}. For baseline methods specifically designed for outage detection with unknown post-outage distribution, we consider an approximated maximum likelihood estimation ({\bf MLE}) proposed to learn the unknown parameters \cite{liao2019structural,liao2021quick}, a generalized likelihood ratio test ({\bf GLRT}) that only considers finite possibilities \cite{chen2015quickest} of post-outage distributions $f$, a {\bf Shewhart} test \cite{rovatsos2016comparison} 
 that utilizes meanshift and covariance changes in the data to detect outages. For methods that are developed for unknown post-change distribution in the change point detection, we consider a non-parametric binned generalized statistic ({\bf BGS}) proposed to approximate the original ratio test in classic CPD \cite{lau2017quickest}, a non-parametric uncertain likelihood ratio ({\bf ULR}) proposed to replace the original ratio \cite{hare2021toward}, a distributed approach ({\bf DIS}) \cite{samudrala2020distributed}, and a deep Q-network approach ({\bf DCQ}) \cite{ma2021deep}.

For more robust evaluation, each experiment will be conducted by the Monte Carlo simulation with over 1000 replications. 
In every replication, we randomly simulate outage time $\lambda$ through geometric distribution $\text{Geo}(\rho)$. This geometric prior is based on our belief that outages can occur independently at any time step, with an equal probability of $\rho$. We choose $\rho=0.04$ in our experiments, which is derived from historical outage data, indicating that each time step has a 4\% chance of experiencing a line outage.

\subsection{Parameters Estimation with Accuracy and Convergence}
\begin{figure}[H]
\begin{center}
\vskip -0.15in
\includegraphics[width=0.8\linewidth]{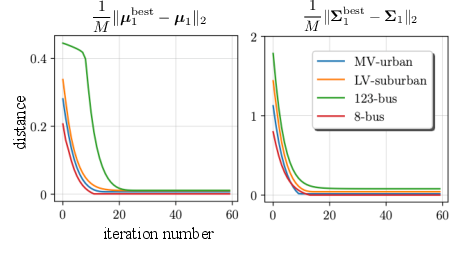}
\vskip -0.15in
\caption{Distance between best update and ground truth against iterations.}
\label{fig:converge}
\end{center}
\end{figure}

Prior to demonstrating the accurate identification of outages with unknown post-outage distribution parameters, we must first verify that our method can learn the optimal parameters with a guaranteed convergence. Throughout the parameter learning iterations, we plot the Euclidean distance between the best update and the ground truth in Fig. \ref{fig:converge}. The plot indicates that our learning process converges to the ground truth, thereby verifying the convergence conclusion stated in Theorem \ref{theorem:mu_converge}.

\vspace{-1.5em}
\subsection{Outage Detection with Small Delay and Rare False Alarm}
After evaluating the effectiveness of using PGD to learn the unknown parameters, we then verify the performance of outage detection using such learned parameters.

\begin{figure}[ht]
\centering
\includegraphics[width=1\linewidth]{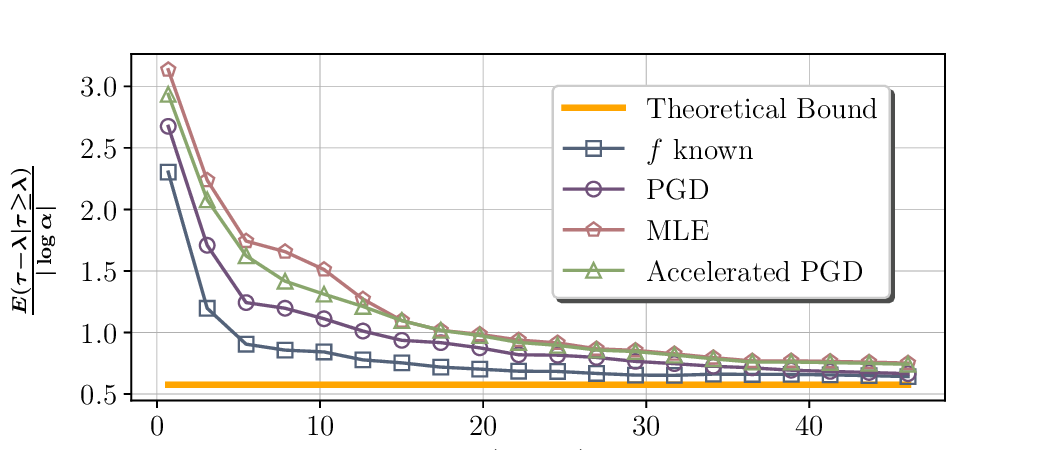}
\caption{Plots of the slope $\frac{1}{|\log\alpha|}\E( \tau-\lambda|\tau\ge\lambda)$ against $|\log\alpha|$ for outage detection in loopy $8$-bus system (outage branch 4-7).}
\label{fig:delay_8}
\end{figure}

The first criterion to evaluate our detection procedure is the average detection delay. To validate the asymptotic optimality of the detection delay in Theorem \ref{theorm:Bayesian}, in Fig. \ref{fig:delay_8}, we plot the average delay $\E(\tau-\lambda|\tau\ge\lambda)$ divided by $|\log\alpha|$ and the theoretical lower bound $-\log(1-\rho)+D_{KL}(f||g)$. We observe that the average detection delay of the case when $f$ is known and that of the PGD both achieve the optimal lower bound asymptotically, while the delay of PGD is slightly higher. Moreover, using PGD and accelerated PGD to learn the unknown post-outage distribution statistics enables quicker line outage detection compared to the method of MLE.

\begin{table*}[ht]
    \centering
    \caption{Performance Comparison on Various Systems, $\alpha=1\%$.}
    \vskip -0.1in
    \begin{tabular}{cc|ccc|cccc|ccc}
    \toprule
    \multicolumn{2}{c}{System} & \multicolumn{3}{c}{System information} & \multicolumn{4}{c}{Average Detection Delay ($1$ unit)} & \multicolumn{3}{c}{Empirical False Alarm Rate $(\%)$} \\
    \midrule
     & \multicolumn{1}{c}{} & \# Branches & \# DERs & Outage  & Optimal & $f$ known & PGD & MLE & $f$ known & PGD & MLE \\ \midrule
        &$8$-bus &  $9$  &$0$ & $4$-$7$  & $2.70$ & $3.13$  & $4.28$  & $5.21$  & $0.9$   & $3.1$  & $44.2$\\
    \multirow{ 3}{*}{Mesh network }&$8$-bus &  $9$  &$8$ & $4$-$7$ & $3.12$  & $3.83$  & $5.02$  & $5.81$  & $1.4$   & $3.3$  & $46.9$\\
    &$123$-bus &  $124$  &$0$ & $73$-$74$ & $0.76$ & $0.91$  & $1.43$  & $1.51$  & $0.5$  & $0.7$  & $52.8$\\
    &$123$-bus &  $124$  &$0$ & $73$-$74$,$14$-$15$ & $0.73$ & $0.87$  & $1.35$  & $1.48$  & $0.4$  & $0.7$  & $51.3$\\
    &$123$-bus &  $124$  &$0$ & $5$ branches & $0.65$ & $0.74$  & $0.99$  & $1.26$  & $0.6$  & $1.1$  & $47.8$\\
    (added loops)&LV suburban & $129$   & $0$  & $26$-$95$& $2.99$ & $3.77$  & $4.65$  & $4.91$  &  $1.5$   & $3.7$  & $53.5$\\
    &LV suburban & $129$   & $30$  & $26$-$95$& $3.24$ & $3.92$  & $4.85$  & $5.14$  &  $1.3$   & $2.4$  & $48.0$\\
    &MV urban & $48$   &$0$   & $34$-$35$ & $0.56$ & $0.78$  & $1.34$   &$1.31$    &$0.3$ & $0.6$ & $46.1$\\
    &MV urban & $48$   &$7$   & $34$-$35$ & $1.01$ & $1.58$  & $2.04$   &$2.33$    &$0.7$ & $1.1$ & $45.9$\\
    \midrule
    &$8$-bus &  $7$  &$8$ & $4$-$7$ & $2.96$  & $3.47$  & $4.51$  & $5.68$  & $0.8$   & $3.3$  & $42.2$\\
    \multirow{ 3}{*}{Radial network }&$8$-bus &  $7$  &$8$ & $2$-$6$  & $3.02$ & $3.60$  & $4.73$  & $5.81$  & $0.8$   & $2.5$  & $45.4$\\
    &$123$-bus &  $122$ &$12$ & $73$-$74$ & $0.98$ & $1.32$  & $1.87$  & $2.32$  & $0.5$  & $1.1$  & $46.5$\\
    &$123$-bus &  $122$ &$122$ & $73$-$74$ & $6.23$ & $6.90$  & $7.73$  & $9.24$  & $2.7$  & $4.9$  & $69.3$\\
    (with DERs)&LV suburban & $114$   & $30$  & $26$-$95$ & $3.15$ & $3.34$  & $4.12$  & $4.35$  &  $1.3$   & $3.2$  & $44.7$\\
    &LV suburban & $114$   & $113$  & $26$-$95$ & $8.62$ & $9.33$  & $10.46$  & $12.70$  &  $3.8$   & $5.4$  & $65.8$\\
    &MV urban & $38$   &$7$   & $34$-$35$ & $0.88$ & $1.45$  & $1.97$   &$2.12$    &$0.6$ & $1.1$ & $49.3$\\
    &MV urban & $38$   &$7$   & $23$-$35$ & $1.02$ & $1.73$  & $2.24$   &$2.49$    &$1.1$ & $1.7$ & $43.0$\\
    \midrule
    \end{tabular}
    \label{tab:my_label}
    \vskip -0.15in
\end{table*}

\begin{figure}[ht]
\centering
\includegraphics[width=1\linewidth]{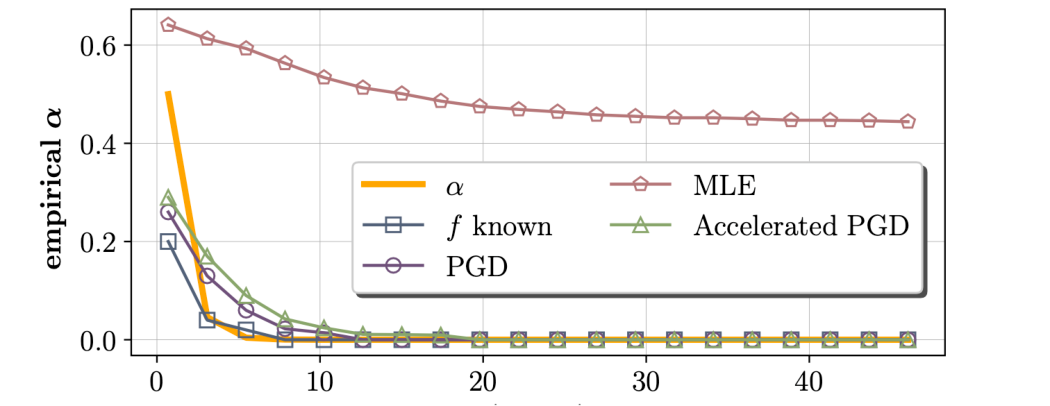}
\vskip -0.15in
\caption{Plots of the empirical false alarm rate against the theoretical probability of false alarm $\alpha$ in loopy $8$-bus system (outage branch 4-7).}
\label{fig:fdr_8}
\end{figure}

The detection rule in Theorem \ref{theorm:Bayesian} can also restrict the false alarm rate below maximum tolerance $\alpha$. To verify this, we calculate the empirical false alarm rate $\P(\tau<\lambda)$ and compare it against the upper bound $\alpha$, as shown in Fig. \ref{fig:fdr_8}. Our proposed method has similar performance compared to the case when $f$ is known since the empirical false alarm is mainly below the upper bound $\alpha$ (especially when $\alpha\to0$). This observation demonstrates that our proposed algorithm could quickly detect line outages with a low false alarm rate, even when the post-outage distribution statistics are unknown.

In Table \ref{tab:my_label}, we present a summary of our proposed method's performance in various grid systems under different outage configurations. Our method demonstrates the ability to handle diverse outage scenarios in both mesh and radial networks with DERs penetration. Specifically, when $f$ is unknown, our method exhibits a lower detection delay and significantly lower false alarm rate in comparison to MLE. Even when $f$ is given, our proposed method only experiences a slight degradation compared to the benchmark. Furthermore, Table \ref{tab:my_label} reveals two additional phenomena. First, when multiple branches are out-of-service simultaneously, the average detection delay is shorter than in a single-line outage scenario due to the larger KL distance between distributions $g$ and $f$ when multiple lines are disconnected. Second, in the radial network with more simulated DERs, it takes more time to detect the line outage as the KL distance between $g$ and $f$ is smaller in this case.

To compare with more relevant methods in the literature, we provide in Table \ref{tab:morecompare} the detection performance of our proposed method and other methods. The comparison of average detection delay and false alarm rate shows that our method is only slightly degraded from the benchmark even though we have incomplete information, and outperforms other methods that also has incomplete information. The reason for this is our performance guarantee, as stated in Theorem \ref{theorem:mu_converge}, which ensures the accurate estimation of unknown post-outage distribution parameters. Furthermore, upon comparing our approach (PGD) with the machine-learning-based method (DCQ), we notice that the latter displays a greater variance in the average detection delay and false alarm rate. This can be attributed to the fact that the neural network's parameters are randomly initialized during training, leading to a more varied estimation of the unknown post-outage distribution parameters.

\begin{table}[H]
\centering
\caption{Comparison with benchmark ($f$ known) and other baseline methods dealing with unknown $f$, $\alpha=1\%$.}
\vskip -0.1in
\begin{tabular}{l|c|c}
    \toprule
    Method & Average Detection Delay & False Alarm Rate(\%) \\
    \midrule
    \midrule
    $f$ known & \multirow{2}{*}{$3.67\pm 0.65$} & \multirow{2}{*}{$0.05\pm 0.03$}\\
     (benchmark) & & \\
    \midrule
    PGD & ${\bf 4.12}\pm 0.83$ & ${\bf 1.06}\pm 0.37$\\
    MLE & $4.56\pm 0.77$ & $38.2\pm 7.39$ \\
    GLRT & $5.03\pm 1.17$ & $7.52\pm 1.01$ \\
    Shewhart & $4.78\pm 0.84$ & $3.84\pm 0.95$ \\
    \midrule
    BGS & $4.85\pm 0.81$ & $4.95\pm 0.81$ \\
    ULR & $4.69\pm 1.01$ & $3.16\pm 0.74$ \\
    DIS & $5.03\pm 0.95$ & $4.28\pm 0.91$ \\
    DCQ & $4.77\pm 1.22$ & $5.16\pm 1.08$ \\
    \bottomrule
\end{tabular}
\label{tab:morecompare}
\end{table}

\vspace{-0.5em}
\subsection{Analysis of Execution Time for Timely Operation}
In addition to detecting delay and false alarm rate, the execution time of the proposed method is also critical for timely detection. Table \ref{tab:time} presents the execution time of Algorithm \ref{algo:PGD} on various grid systems with different sampling rates.
\begin{table}[H]
    \centering
    \vskip -0.15in
    \caption{Executing time (second) of algorithm \ref{algo:PGD}.}
    \vskip -0.1in
    \begin{tabular}{c|c|cc}
    \toprule
    System (resolution) & $f$-known & PGD & Accelerated PGD \\
    \midrule
    $8$-bus (1 hour)        & $0.138$    & $1.029$ & $0.468$\\
    $123$-bus (1 hour)      & $0.453$    & $2.019$ & $1.176$ \\
    $8$-bus (15 minute)     & $0.234$    & $1.263$ & $0.582$\\
    $123$-bus (15 minute)   & $0.579$    & $2.671$ & $1.236$ \\
    \midrule
    \end{tabular}
    \label{tab:time}
    \vskip -0.10in
\end{table}

From the records, less than $3$ seconds per sample is needed to obtain the outage detection result when we receive a new sample, even for grid systems with more than $100$ buses. This execution time can be negligible compared to the normal smart meter sampling interval, which ranges from $1$ minute to $1$ hour. Moreover, since the most time-consuming part of our algorithm is the matrix exponential and matrix logarithm operation, we can accelerate the algorithm by approximating these operations based on their Taylor series expansion, as discussed in Section \ref{sec:speedup}. 

\begin{figure}[ht]
\centering
\vskip -0.10in
\includegraphics[width=1\linewidth]{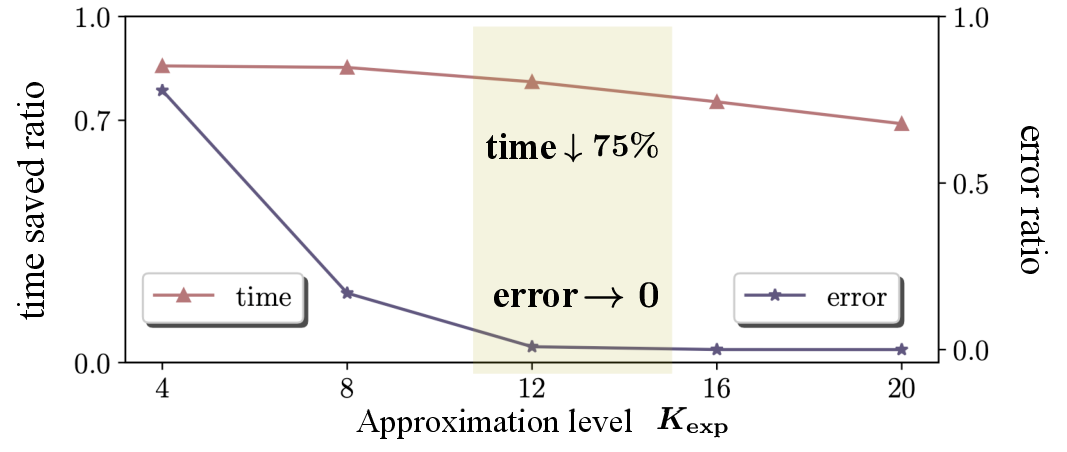}
\vskip -0.1in
\caption{The ratio of saved execution time versus the ratio of error incurred by the operation $\widehat{\exp}$ against the level of approximation $\Lexp$.}
\vskip -0.10in
\label{fig:speed_up_exp}
\end{figure}

To maintain the detection performance, we select an appropriate level of approximation with near-zero errors incurred. In Fig. \ref{fig:speed_up_exp}, we choose $\Lexp=12$ because at this approximation level, the executing time is reduced by more than $75\%$ with almost zero errors incurred. Similarly, we choose $\Llog=16$ which is slightly larger than $\Lexp$ since the term $\frac{1}{k}$ in \eqref{eq:log_app} decreases slower than term $\frac{1}{k!}$ in \eqref{eq:exp_app} as $k$ becomes large. As a result, the accelerated PGD only shows a slight performance degradation, as shown in Fig. \ref{fig:delay_8} and \ref{fig:fdr_8}. More importantly, in Table \ref{tab:time}, the acceleration technique reduces the execution time by more than half, thus achieving better timely outage detection. Evidently, the acceleration technique becomes more valuable in distribution systems with smart meters of a lower sampling rate, which is the trend of the future.

Table \ref{tab:time} exhibits another phenomenon: as the sampling rate increases, the processing time for accumulated data $\mv^{1:N}$ also increases. Consequently, conducting Algorithm \ref{algo:PGD} becomes challenging when $N$ grows very large. To address this issue, we discovered that a small window of historical samples can adequately differentiate between the pre- and post-outage distributions. Specifically, instead of using all $N$ samples when $N$ is very large, we can employ the latest $N_0$ samples to represent the entire data stream since they contain nearly identical distribution information in the temporal dimension. By doing so, the time complexity of the algorithm is restricted to a constant number, $N_0$. Through experiments, we determined that $N_0=100$ samples are sufficient to maintain the algorithm's effectiveness and accuracy.

\vspace{-1em}
\subsection{Outage Branch Localization with Accuracy}
\label{sec:local_experiment}
\begin{figure}[ht]
\centering
\vskip -0.15in
\includegraphics[width=1\linewidth]{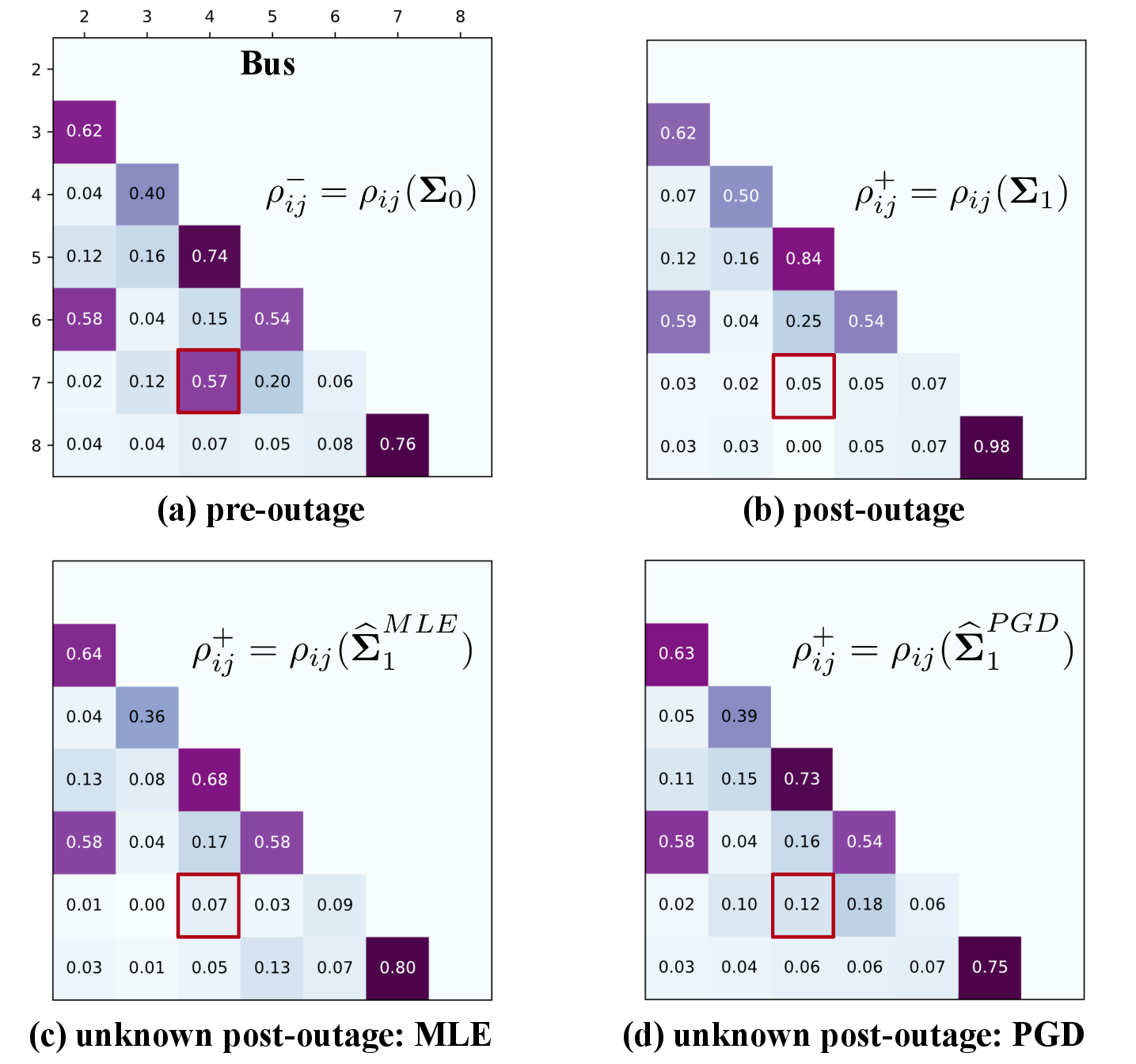}
\vskip -0.1in
\caption{Absolute conditional correlation of the loopy 8-bus system before and after an outage in branches 4-7. We choose $\delta_{\max}=0.5$ and $\delta_{\min}=0.1$.}
\vskip -0.1in
\label{fig:localization}
\end{figure}

After detecting an outage occurrence, we further compute the conditional correlation between buses to localize the out-of-service branch, following Theorem \ref{lemma:local}. 
Here, Fig. \ref{fig:localization} demonstrates the absolute conditional correlation of every pair of buses in the loopy 8-bus system before and after a line outage at branch 4-7. Since the value in the red box changes from a non-zero value before the outage ($\rho^{-}_{47}>\delta_{\max}$) to near zero after the outage ($\rho^{+}_{47}<\delta_{\min}$), we localize the out-of-service branch at 4-7, which matches the ground truth.
Fig. \ref{fig:localization}(d) indicates that the localization method using the learned covariance matrix through PGD is as effective as the optimal scenario, and is more effective than using the learned covariance matrix through MLE. 

Table \ref{tab:localize} demonstrates the accuracy rate of localization in 1,000 experiments. As shown, our proposed method can accurately localize over 90\% of the outage branches, even without the post-outage distribution parameters.

\begin{table}[H]
    \centering
    \vskip -0.15in
    \caption{Localization accuracy (\%) in algorithm \ref{algo:PGD}.}
    \vskip -0.1in
    \begin{tabular}{c|c|cc}
    \toprule
    System & $f$-known & MLE & PGD \\
    \midrule
    $8$-bus         & $98.3$    & $92.5$ & $93.6$\\
    $123$-bus       & $94.7$    & $88.2$ & $91.8$ \\
    LV suburban     & $95.6$    & $90.9$ & $92.3$\\
    MV urban        & $95.1$    & $92.3$ & $94.6$ \\
    \midrule
    \end{tabular}
    \label{tab:localize}
    \vskip -0.2in
\end{table}

\subsection{Sensitivity Analysis to Data Noise and Data Coverage}
\label{sec:sensitivity}
Smart-meter data can be noisy and corrupted. 
Besides, smart-meter data may not be accessible in every household of the distribution grid. Thus, an analysis of our proposed method under different levels of data noise and data coverage is critical to gain a better understanding of its effectiveness in real-world outage scenarios.

In the U.S., ANSI C12.20 standard permits the utility smart meters to have an error within $\pm 0.5\%$ \cite{zheng2013smart}. Hence, we simulate such noise in our smart-meter voltage measurements and then evaluate the corresponding detection performance. Table \ref{tab:noise} shows both average detection delay and false alarm rate under our method with different noise levels. As we see, when the noise level is $0.5\%$, one more sample (compared to noiseless case) is needed for the detection, while the false alarm rate is also slightly increased. In fact, we are able to quantify the increase in detection delay by analyzing the change of KL divergence between the pre- and post-outage distributions affected by noisy data. In doing so, we are able to better understand and control real-world line outage detection.

\begin{table}[H]
    \centering
    \vskip -0.15in
    \caption{Outage detection performance with noisy data, $\alpha=1\%$.}
    \vskip -0.1in
    \begin{tabular}{c|c|c}
    \toprule
    Noise level & Average Detection Delay & False Alarm Rate \\
    \midrule
    noiseless  & $2.51$ & $1.26\%$\\
    0.1\%       & $2.76$ & $1.36\%$\\
    0.5\%       & $3.64$ & $2.88\%$ \\
    \midrule
    \end{tabular}
    \label{tab:noise}
    \vskip -0.15in
\end{table}

Another concern regarding the smart meter data is that it may not be accessible for every household in the distribution grid, particularly in certain situations. For instance, (1) in rural areas, some households may not have installed smart meters, (2) the voltage data for certain households may be lost due to technical issues, and (3) some households may refuse to provide their voltage data due to privacy concerns. Although the new generation of smart meters is developing very fast, an analysis of incomplete coverage of smart meters data is needed to evaluate our algorithm. We first emphasize that our proposed method does not rely on the assumption of 100\% coverage of smart meters data in the grid. In fact, a power line outage will influence almost all buses in the system, while the degree of influence depends on the distance between a bus and the source of the outage. Hence, we can reveal the outage by detecting the distribution change of some (not necessarily all) voltage data collected nearby the outage source. 

\begin{figure}[H]
\centering
\vskip -0.10in
\includegraphics[width=0.9\linewidth]{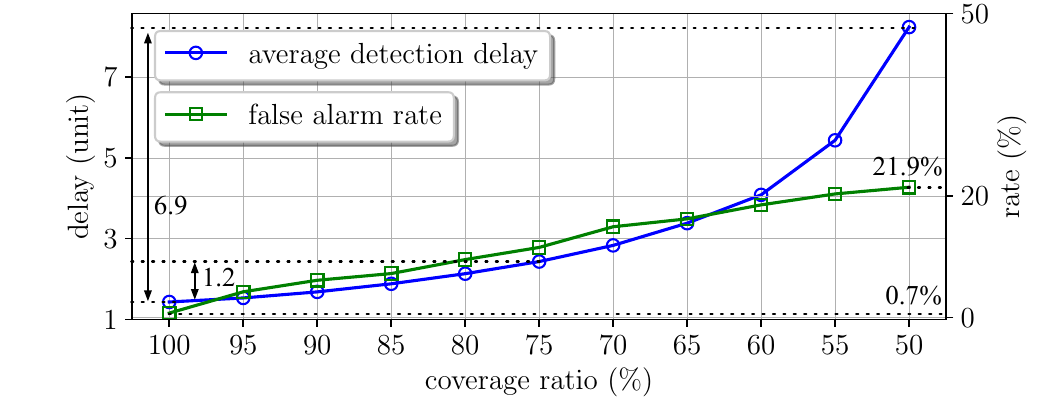}
\vskip -0.15in
\caption{Average detection delay (unit) and false alarm rate (\%) under different levels of data coverage in loopy 123-bus system, $\alpha=1\%$.}
\vskip -0.10in
\label{fig:coverage}
\end{figure}

According to \cite{USreport2}, over 107 million smart meters were deployed by 2021, covering 75\% of U.S. households. Hence, we simulate this scenario where only a ratio of buses is randomly selected to provide its voltage measurements in the grid system to detect the outage. Fig. \ref{fig:coverage} demonstrates both the average detection delay and the false alarm rate of our method under different levels of coverage ratio.
In comparison to the scenario where voltage data is available for all buses, the detection delay increases by 1.2 units of time step. This means that an extra 1.2 samples of data are needed to detect the outage in the 75\% data coverage scenario. Similarly, when the data coverage ratio drops to 50\%, an additional 6.9 samples are required for detection. Furthermore, as the data coverage ratio decreases to only 50\%, the false alarm rate increases from 0.7\% to 21.9\%.

\vspace{-0.5em}
\subsection{Sensitivity Analysis to Hyper-parameters}
Our detection procedure involves certain hyper-parameters that have the potential to influence the detection performance, such as the geometric distribution parameter $\rho$. Therefore, conducting a sensitivity analysis pertaining to these hyper-parameters is crucial to assess the robustness of our proposed method.

\begin{figure}[H]
\centering
\vskip -0.10in
\includegraphics[width=0.9\linewidth]{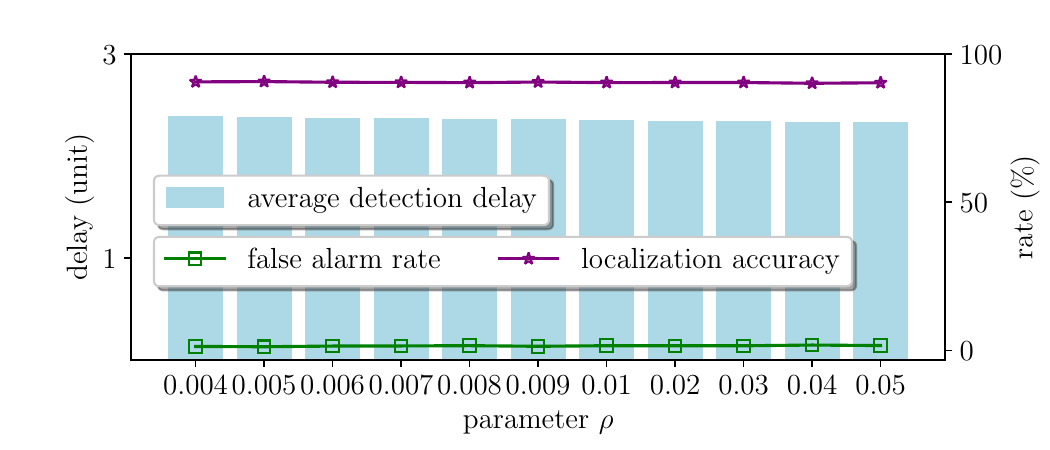}
\vskip -0.15in
\caption{Average detection delay (unit), false alarm rate (\%) and localization accuracy (\%) under different levels of $\rho$ in loopy 123-bus system, $\alpha=1\%$.}
\label{fig:rho}
\end{figure}

During our experiments, we randomly simulated the outage time $\lambda$ using a geometric prior distribution denoted as $\text{Geo}(\rho)$. This distribution aligns with our assumption that outages can take place in any time step with an equal probability of $\rho$. Fig. \ref{fig:rho} illustrates the effect of the parameter $\rho$ on the performance of our detection method. It can be observed that choosing different values of $\rho$ within the range of $0.004$ to $0.05$ has a negligible impact on both the false alarm rate (approximately 1.65\%) and the localization accuracy (approximately 92.8\%). Additionally, decreasing the value of $\rho$ leads to a slight increase in the average detection delay.

\section{Limitations}
While this paper has some performance guarantee, we also encounter some of the limitations that we look forward to address in the future. For instance, while the proposed approach requires only voltage magnitude data, it may be limited by the quality and availability of this data. As shown in Section \ref{sec:sensitivity}, noise or incomplete data will lead to additional detection delay. Future research could investigate how to leverage additional types of data to improve outage detection and localization. Another aspect worth investigating is the ability to withstand diverse outage scenarios. For instance, if an outage occurs in an insignificant branch of the grid, resulting in minimal fluctuations in voltage data, detecting such subtle outages remains a challenge. Hence, further research is necessary to improve the detection performance in such cases. Lastly, although sensor readings facilitate line outage detection, they pose privacy concerns since they can disclose sensitive information like household occupancy and economic status to potential adversaries. An open problem is how to identify outages accurately without compromising the customer's data.

\section{Conclusion}
\label{sec:conclusion}
This paper resolves three challenges in the line outage identification problem: data availability, unknown outage pattern, and timely operation.
Our approach for detecting and localizing line outages only utilizes voltage magnitude. To handle unknown outage patterns, we propose a Projected Gradient Descent framework that can learn the unknown post-outage distribution parameters with a feasibility guarantee. We demonstrate the convergence guarantee of our method and further accelerate the proposed algorithm for timely operation, resulting in a reduction of more than 75\% of execution time with minimal errors. Empirical results on representative grid systems confirm that our proposed method is suitable for timely outage detection and localization, even in the absence of prior knowledge about outage patterns.

\vspace{-0.8em}
\appendices

\section{Proof of Lemma \ref{lemma:admit}}
\begin{proof}
In \eqref{eq:admit}, the incidence matrix $\incidence$ has rank $M-1$ in the connected grid \cite{bapat2010graphs}. The diagonal and full-rank complex matrix $\admitt$ can be decomposed as $\admitt=\boldsymbol{B}^\top\boldsymbol{B}$ where $\boldsymbol{B}$ is also full-rank. Then, $\rank(\incidence^\top \admitt \incidence)=M-1$ since
$$
\incidence^\top \admitt \incidence = \incidence^\top \boldsymbol{B}^\top\boldsymbol{B} \incidence = (\boldsymbol{B} \incidence)^\top(\boldsymbol{B} \incidence).
$$
Hence, we have $\rank(\admit)=M-1$ considering zero shunt admittance and $\rank(\admit)=M$ otherwise. In both cases, $\admit$ has at least one non-singular ${(M-1)\times(M-1)}$ sub-matrix, which can be viewed as the remaining matrix after eliminating one column and one row (of a slack-bus) in $\admit$. \qedhere
\end{proof}

\section{Proof of Theorem \ref{theorem:mu_converge}}
\label{app:proofconvergence}
\begin{proof}
Since $\lfun(\postmu)$ is restricted convex, we apply this convexity property in constraint set $\mathcal{U}$ to give an upper bound to the level of sub-optimality of the $e^{th}$ iterate as
\begin{equation}\label{eq:mu_objective_gap}
\lfun_e = \lfun(\postmu\eth) - \lfun(\postmu^{\ast}) \le \langle \nabla \lfun(\postmu\eth), \postmu\eth - \postmu^{\ast} \rangle.
\end{equation}
We can obtain an upper bound of $\langle \nabla \lfun(\postmu\eth), \postmu\eth - \postmu^{\ast} \rangle$ as
\begin{align}\label{eq:mu_objective_scale}
    &\frac{1}{2\eta}\left(\|\postmu\eth - \postmu^{\ast}\|_2^2 + \eta^2 U^2 - \|\postmu^{e+1} - \postmu^{\ast}\|_2^2 \right),
\end{align}
where the inequality holds since the gradient $\nabla \lfun(\postmu)$ 
can be bounded on constraint set $\mathcal{U}$, i.e., $\|\nabla \lfun(\postmu)\|_2 \le U$ for all $\postmu\in\mathcal{U}$. Combining equations (\ref{eq:mu_objective_gap}) and (\ref{eq:mu_objective_scale}), we arrive at
\begin{equation*}
    \lfun_e \le \frac{1}{2\eta}\left(\|\postmu\eth - \postmu^{\ast}\|_2^2 - \|\postmu^{e+1} - \postmu^{\ast}\|_2^2 \right) + \frac{\eta U^2}{2},
\end{equation*}
which upper bounds the sub-optimality at every $e^{th}$ iterate. Suppose we initialize the mean vector as $\postmu^1$, we can sum the sub-optimality across iterates and average it by dividing the total iterate number $E$ as
\vspace{-0.1em}
$
    \frac{1}{E}\sum_{e=1}^E\lfun_e \le \frac{1}{2\sqrt{E}} \left(\|\postmu^{\ast}-\postmu^1\|_2^2 + U^2 \right) ,
$
where step size $\eta = \frac{1}{\sqrt{E}}$. Therefore, for any $\varepsilon>0$, we can always use at most $E = \mathcal{O}(\frac{1}{\varepsilon^2})$ total iterates to make sure $\frac{1}{E}\sum_{e=1}^E\lfun_e \le \varepsilon$ since $\|\postmu^{\ast}-\postmu^1\|_2^2 + U^2$ is a constant number.

Then, we prove that the averaged and best update both converge to $\postmu^{\ast}$ after $E$ updates. Applying Jensen's inequality, we derive
$
    \lfun(\postmu^{\text{avg}}) = \lfun(\frac{1}{E}\sum_{e=1}^E \postmu\eth) \le \frac{1}{E}\sum_{e=1}^E \lfun(\postmu\eth) \le \lfun(\postmu^{\ast}) + \varepsilon.
$
Since $\lfun(\postmu^{\text{best}}) \le \lfun(\postmu\eth)$ for every $e^{th}$ iterate, we have
$
    \lfun(\postmu^{\text{best}}) \le \frac{1}{E}\sum_{e=1}^E \lfun(\postmu\eth) \le \lfun(\postmu^{\ast}) + \varepsilon.\qquad\qquad \qedhere
$
\end{proof}

\vspace{-1em}
\bibliographystyle{IEEEtran}
\bibliography{ref}

\end{document}